\documentclass[conference]{IEEEtran}
\usepackage{times}
\usepackage[numbers]{natbib}
\usepackage{multicol}

\usepackage{amsmath,amssymb}
\usepackage{booktabs}
\usepackage{array}
\usepackage{multirow}
\usepackage{graphicx}
\usepackage{capt-of}   %
\usepackage{xcolor}
\usepackage{colortbl}
\usepackage{tikz}
\usepackage{fontawesome5}
\usepackage{pbalance}
\usetikzlibrary{arrows.meta,positioning,calc,decorations.pathreplacing}
\usepackage[bookmarks=true,hidelinks]{hyperref}

\newcommand{\X}{\mathbf{X}}

\definecolor{lane-s}{RGB}{74,85,104}
\definecolor{lane-g}{RGB}{31,111,139}
\definecolor{lane-p}{RGB}{140,88,32}
\definecolor{fill-s}{RGB}{231,234,239}
\definecolor{fill-g}{RGB}{222,237,242}
\definecolor{fill-p}{RGB}{246,231,220}

\definecolor{fill-ours}{RGB}{226,238,214}
\newcommand{\ours}{\rowcolor{fill-ours}}
\newcommand{\best}[1]{\textbf{#1}}
\newcommand{\cat}[2]{\multicolumn{#1}{l}{\textit{#2}}}

\begin{document}

\makeatletter
\let\@oldmaketitle\@maketitle
\renewcommand{\@maketitle}{\@oldmaketitle\begin{center}
\resizebox{\textwidth}{!}{%
\begin{tikzpicture}[
  font=\sffamily\small,
  >=Stealth,
  sub/.style={draw=lane-g, fill=white, rounded corners=2pt},
  link/.style={line width=4.5pt, line cap=round, draw=lane-s!85},
  jnt/.style={circle, fill=white, draw=lane-s, line width=1.2pt, inner sep=2.6pt},
  flow/.style={->, line width=1pt, draw=black!70},
  big/.style={-{Triangle[length=2.8mm, width=5.6mm]}, line width=2.6mm, draw=black!40},
  lab/.style={font=\sffamily\footnotesize, text=black!70},
  slab/.style={font=\sffamily\scriptsize, text=black!70},
  mod/.style={font=\sffamily\small\bfseries},                  %
  smod/.style={font=\sffamily\footnotesize\bfseries},          %
  fire/.style={text=orange!90!red},
  axx/.style={->, line width=1.0pt, draw=red!75!black},
  axy/.style={->, line width=1.0pt, draw=green!50!black},
  axz/.style={->, line width=1.0pt, draw=blue!75!black},
  ring/.style={line width=1.3pt},
]
\colorlet{c-hand}{lane-s}
\colorlet{c-bowl}{lane-g}
\colorlet{c-plate}{teal!60!black}
\colorlet{c-bottle}{violet!75!black}
\colorlet{c-cab}{orange!75!black}
\colorlet{c-stove}{red!65!black}
\definecolor{c-ice}{RGB}{120,195,235}
\colorlet{c-fire}{orange!85!yellow}
\definecolor{c-zero}{RGB}{70,130,40}

\newcommand{\modulebox}[7]{%
  \fill[#6, rounded corners=3pt] (#1,#2) rectangle (#3,#4);
  \begin{scope}
    \clip[rounded corners=3pt] (#1,#2) rectangle (#3,#4);
    \fill[#5] (#1,#4-0.5) rectangle (#3,#4);
  \end{scope}
  \draw[#5, line width=0.9pt, rounded corners=3pt] (#1,#2) rectangle (#3,#4);
  \node[mod, text=white] at ({(#1+#3)/2},#4-0.25) {#7};}
\newcommand{\neuron}[2]{%
  \draw[lane-s, fill=white, line width=0.8pt] (#1,#2) circle (0.075);}
\newcommand{\actneuron}[2]{%
  \draw[lane-p, fill=lane-p!55, line width=0.8pt] (#1,#2) circle (0.075);}
\newcommand{\boxat}[3]{%
  \begin{scope}[shift={(#1,#2)}, scale=#3]
    \draw[lane-g, line width=0.7pt] (-0.16,-0.13) rectangle (0.10,0.11);
    \draw[lane-g, line width=0.7pt] (-0.16,0.11) -- (-0.08,0.19) -- (0.18,0.19) -- (0.10,0.11);
    \draw[lane-g, line width=0.7pt] (0.18,0.19) -- (0.18,-0.05) -- (0.10,-0.13);
  \end{scope}}
\newcommand{\screwat}[4]{%
  \begin{scope}[shift={(#1,#2)}, rotate=#3, scale=#4]
    \draw[lane-g, line width=1.1pt, -{Stealth[length=1.6mm]}]
         (-0.42,0) -- (0.50,0);
    \draw[lane-g, line width=0.9pt, domain=0:15.71, samples=120, smooth,
          variable=\t]
         plot ({-0.34+0.043*\t+0.06*sin(\t r)}, {0.17*cos(\t r)});
  \end{scope}}
\newcommand{\tokenat}[4]{%
  \begin{scope}[opacity=#4]
  \draw[thick, draw=#3, fill=fill-g, rounded corners=1.5pt]
       (#1-0.5,-1.55) rectangle (#1+0.5,-0.95);
  \screwat{#1-0.12}{-1.25}{0}{0.72}
  \boxat{#1+0.34}{-1.27}{0.9}
  \draw[thick, draw=#3, fill=fill-s, rounded corners=1.5pt]
       (#1-0.5,-2.15) rectangle (#1+0.5,-1.55);
  \neuron{#1-0.25}{-1.85} \neuron{#1}{-1.85} \actneuron{#1+0.25}{-1.85}
  \node[slab] at (#1,-2.4) {#2};
  \end{scope}}
\newcommand{\detected}[3]{%
  \node[inner sep=0] at (#1,#2) {#3};
  \draw[lane-g, densely dashed, line width=0.6pt] (#1-0.46,#2-0.42) rectangle (#1+0.46,#2+0.40);
  \draw[axx] (#1,#2-0.1) -- ++(0.30,-0.08);
  \draw[axy] (#1,#2-0.1) -- ++(0.16,0.13);
  \draw[axz] (#1,#2-0.1) -- ++(0,0.30);}

\modulebox{-0.4}{-4.05}{4.07}{1.95}{lane-s}{fill-s!45}{Scene}
\begin{scope}[shift={(-0.115,-2.974)}, scale=0.957]
\node[anchor=south west, inner sep=0] (img) at (0,0)
  {\includegraphics[width=3.9cm]{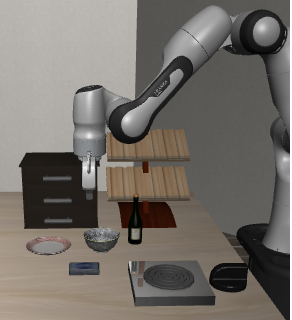}};
\draw[ring, draw=c-hand]  (1.242,2.037) ellipse (0.209 and 0.363);
\draw[ring, draw=c-bowl]  (1.423,1.130) ellipse (0.279 and 0.181);
\draw[ring, draw=c-plate] (0.656,1.053) ellipse (0.349 and 0.160);
\begin{scope}[opacity=0.55]
\draw[ring, draw=c-bottle] (1.884,1.395) ellipse (0.126 and 0.377);
\draw[ring, draw=c-cab, rounded corners=2pt]   (0.279,1.242) rectangle (1.381,2.344);
\draw[ring, draw=c-stove, rounded corners=2pt] (1.772,0.195) rectangle (3.000,0.851);
\end{scope}
\end{scope}
\node[anchor=north west, fill=white, fill opacity=0.85, text opacity=1,
      rounded corners=1.5pt, inner sep=2.5pt, smod]
  at (-0.05,1.25) {Training: simulation};
\node[draw=black!40, rounded corners=2pt, font=\sffamily\footnotesize\itshape,
      inner sep=4pt] (instr) at (1.835,-3.5) {``Put the bowl on the plate''};

\begin{scope}[xshift=-0.53cm]
\modulebox{5.1}{-4.05}{9.6}{1.95}{lane-g}{fill-g}{Scene analyser~\textcolor{c-ice}{\faSnowflake}~{\mdseries frozen}}
\draw[sub] (5.3,-1.85) rectangle (9.4,1.3);
\node[smod, text=lane-g] at (7.35,1.02) {Pose and box detection};
\detected{5.95}{0.30}{\includegraphics[width=1.18cm]{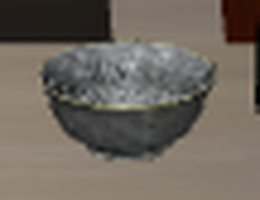}}
\detected{7.35}{0.30}{\includegraphics[width=1.18cm]{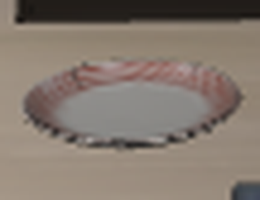}}
\detected{8.75}{0.30}{\includegraphics[width=1.18cm]{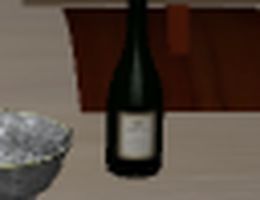}}
\detected{6.65}{-1.08}{\includegraphics[width=1.18cm]{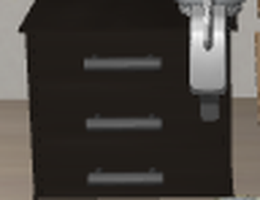}}
\detected{8.05}{-1.08}{\includegraphics[width=1.18cm]{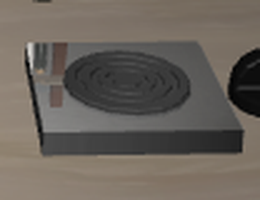}}
\node[slab] at (5.95,-0.32) {Bowl};
\node[slab] at (7.35,-0.32) {Plate};
\node[slab] at (8.75,-0.32) {Bottle};
\node[slab] at (6.65,-1.70) {Cabinet};
\node[slab] at (8.05,-1.70) {Stove};
\draw[sub] (5.3,-3.85) rectangle (9.4,-2.05);
\node[smod, text=lane-g] at (7.35,-2.35) {Language model parser~\faCommentDots};
\node[lab, font=\sffamily\footnotesize\itshape] at (7.35,-2.88) {On(bowl, plate)};
\node[slab] at (7.35,-3.33) {Source: bowl,\enspace target: plate};
\end{scope}

\begin{scope}[xshift=-1.63cm]
\modulebox{11.2}{-4.05}{20.9}{1.95}{lane-s!70!black}{black!3}
  {Screw Attention Policy~\textcolor{c-fire}{\faFire}~{\mdseries trained}}
\node[smod, anchor=west] at (11.4,0.72) {Attention layer $\times 2$};
\draw[draw=black!35, fill=white, rounded corners=2pt] (15.3,0.45) rectangle (20.7,1.25);
\screwat{15.7}{1.04}{0}{0.6}
\boxat{16.15}{1.02}{0.75}
\node[slab, anchor=west] at (16.4,1.04) {Geometric channels $\mathbf{g}_i$ (screws, box)};
\neuron{15.7}{0.65}
\node[slab, anchor=west] at (16.1,0.65) {Scalars $\mathbf{s}_i$};
\actneuron{17.9}{0.65}
\node[slab, anchor=west] at (18.2,0.65) {Action primitive};
\draw[draw=black!55, fill=white, rounded corners=2pt] (11.4,-0.75) rectangle (20.7,0.35);
\draw[black!25] (16.2,-0.65) -- (16.2,0.25);
\node[font=\sffamily\small, text=black!75, anchor=west, align=left] at (11.55,-0.2) {Message\\passing};
\draw[->, line width=1.3pt, draw=lane-p, shorten <=2pt, shorten >=2pt]
  (14.87,-0.7) to[bend right=60] (13.75,-0.7);
\draw[->, line width=1.3pt, draw=lane-p, shorten <=2pt, shorten >=2pt]
  (15.99,-0.7) to[bend right=60] (14.87,-0.7);
\screwat{14.31}{-0.31}{0}{0.55}
\screwat{15.43}{-0.31}{0}{0.55}
\node[font=\sffamily\small, text=lane-p] at (14.87,0.1) {$\mathbf{X}_{ij}$};
\node[font=\sffamily\footnotesize, text=black!75, align=center] at (17.85,-0.2)
  {Scores $\alpha_{ij}$:\\scalar features};
\foreach \row [count=\r from 0] in {{40,75,60,0,0,0},{60,35,70,0,0,0},
    {35,65,40,0,0,0},{25,50,35,0,0,0},{35,45,30,0,0,0},
    {20,55,45,0,0,0}} {%
  \foreach \v [count=\c from 0] in \row {%
    \fill[lane-s!\v] (19.75+0.1*\c,0.05-0.1*\r) rectangle ++(0.1,-0.1);
  }}
\draw[white, line width=0.3pt, step=0.1, shift={(19.75,-0.55)}] (0,0) grid (0.6,0.6);
\draw[black!45, line width=0.4pt] (19.75,0.05) rectangle (20.35,-0.55);
\node[font=\sffamily\scriptsize\itshape, text=black!70] at (19.64,-0.25) {$i$};
\node[font=\sffamily\scriptsize\itshape, text=black!70] at (20.05,0.17) {$j$};
\draw[thick, dashed, draw=lane-s, fill=white, rounded corners=2pt] (11.4,-2.15) rectangle (12.5,-0.95);
\fill[lane-s!60] (11.57,-1.52) rectangle (11.81,-1.45);
\draw[line width=1.6pt, line cap=round, draw=lane-s]
  (11.69,-1.45) -- (11.73,-1.12) -- (12.15,-1.05) -- (12.29,-1.3);
\foreach \p in {(11.73,-1.12),(12.15,-1.05)}
  \fill[white] \p circle (0.035);
\draw[line width=1.1pt, draw=lane-s] (12.24,-1.3) -- (12.24,-1.4) (12.34,-1.3) -- (12.34,-1.4);
\node[font=\sffamily\scriptsize\bfseries, align=center] at (11.95,-1.83) {Robot\\[-2pt]model};
\draw[flow, draw=lane-s] (12.5,-1.25) -- (13.25,-1.25);
\tokenat{13.75}{\shortstack{End\\[-1pt]effector}}{c-hand}{1}
\tokenat{14.87}{Bowl}{c-bowl}{1}
\tokenat{15.99}{Plate}{c-plate}{1}
\tokenat{17.11}{Bottle}{c-bottle}{0.3}
\tokenat{18.23}{Cabinet}{c-cab}{0.3}
\tokenat{19.35}{Stove}{c-stove}{0.3}

\draw[draw=lane-p, fill=fill-p, rounded corners=2pt] (11.4,-3.9) rectangle (20.7,-3.15);
\node[smod] at (16.05,-3.38) {Action primitive estimator};
\node[slab] at (16.05,-3.7) {Reach, grasp, \underline{carry}, place};
\draw[flow, draw=lane-p] (12.85,-3.15) -- (12.85,-1.5) -- (13.25,-1.5);   %
\node[slab, align=right, anchor=east] at (12.79,-2.72) {Grasp\\[-2pt]orientation};
\draw[thick, draw=lane-s] (13.65,-2.72) -- (13.65,-2.82) -- (16.49,-2.82) -- (16.49,-2.72);
\draw[flow, draw=lane-s] (14.87,-2.82) -- (14.87,-3.15);
\node[slab, anchor=west] at (14.95,-2.98) {State};
\draw[thick, densely dashed, draw=lane-p] (16.61,-2.72) -- (16.61,-2.82) -- (19.85,-2.82) -- (19.85,-2.72);
\draw[flow, draw=lane-p, densely dashed] (18.23,-3.15) -- (18.23,-2.82);
\node[slab, anchor=west] at (18.31,-2.98) {Masking};
\draw[flow, draw=lane-p] (20.40,-3.15) -- (20.40,-1.85) -- (19.90,-1.85);
\node[slab, anchor=south west, inner sep=1pt] at (19.92,-1.78) {Action};
\end{scope}

\begin{scope}[xshift=-1.59cm]   %
\modulebox{21.36}{-4.05}{25.83}{1.95}{lane-s}{fill-s!45}{Deployment}
\begin{scope}[shift={(21.645,-1.284)}]
\node[anchor=south west, inner sep=0] at (0,0)
  {\includegraphics[width=3.9cm]{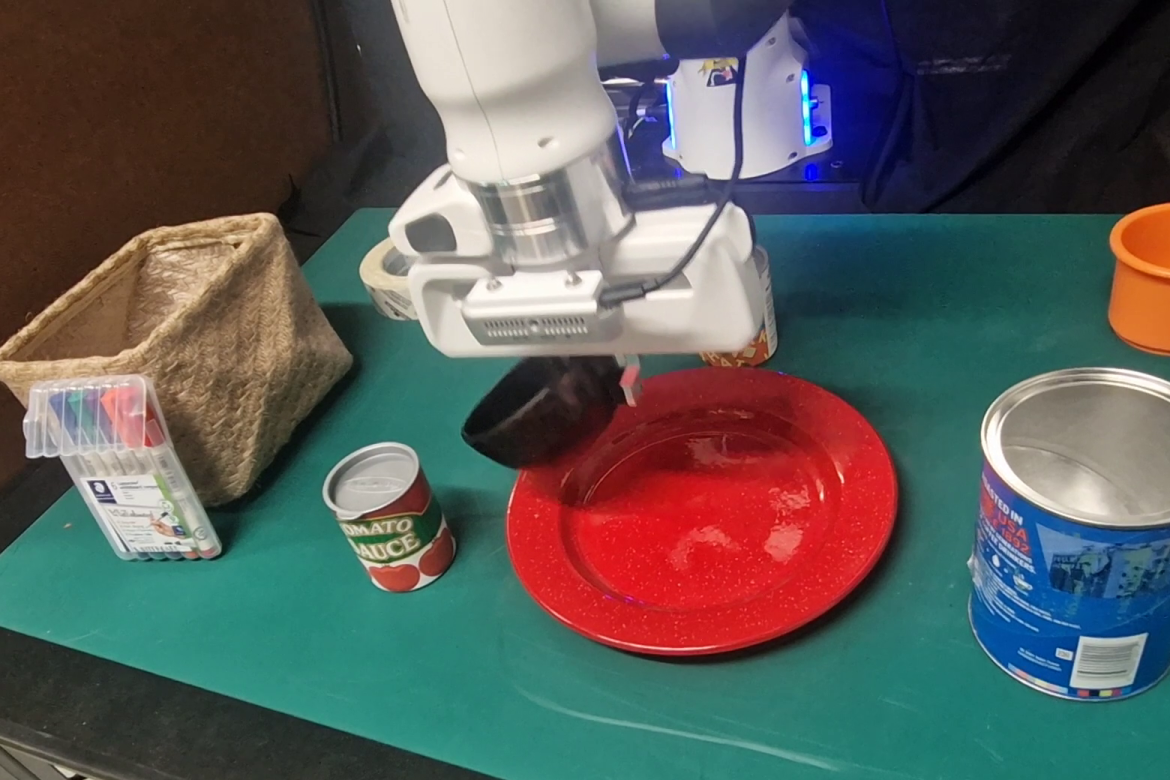}};
\draw[ring, draw=c-hand]  (1.900,1.700) ellipse (0.575 and 0.375);
\draw[ring, draw=c-bowl]  (1.760,1.238) ellipse (0.275 and 0.188);
\draw[ring, draw=c-plate] (2.340,0.888) ellipse (0.650 and 0.488);
\begin{scope}[opacity=0.55]
\draw[ring, draw=c-bottle, rounded corners=2pt] (0.050,0.920) rectangle (1.150,1.880);
\draw[ring, draw=c-cab, rounded corners=2pt]    (0.083,0.745) rectangle (0.730,1.350);
\draw[ring, draw=c-stove]  (1.293,0.883) ellipse (0.218 and 0.250);
\draw[ring, draw=c-cab]    (3.565,0.825) ellipse (0.335 and 0.555);
\end{scope}
\end{scope}
\node[font=\sffamily\scriptsize\bfseries, text=black!75] at (23.595,-2.47)
  {Cross-embodiment, 5 unseen arms};
\foreach \y/\name/\v/\col in {-2.85/{\textbf{Ours}}/80.2/c-zero,
                              -3.21/SkelWAM/33.0/black!35,
                              -3.57/Best VLA/0.2/black!35} {
  \node[anchor=east, font=\sffamily\scriptsize] at (22.90,\y) {\name};
  \fill[\col] (22.95,\y-0.12) rectangle ++({max(0.02,\v/50)},0.24);
  \node[anchor=west, font=\sffamily\scriptsize] at ({22.97+max(0.02,\v/50)},\y) {\v\%};
}
\draw[black!50, line width=0.4pt] (22.95,-3.74) -- (22.95,-2.68);
\node[anchor=north west, fill=white, fill opacity=0.85, text opacity=1,
      rounded corners=1.5pt, inner sep=2.5pt, smod]
  at (21.71,1.25) {Real world};
\node[fill=c-zero, text=white, rounded corners=2pt, anchor=north, align=center,
      font=\sffamily\small\bfseries, minimum width=3.9cm,
      minimum height=0.78cm, inner sep=0pt] at (23.595,-1.37)
  {Zero-shot\\[-2pt]{\scriptsize No real-robot data}};
\end{scope}

\draw[big] (3.95,-1.05) -- (4.69,-1.05);
\draw[big] (8.95,-1.05) -- (9.69,-1.05);
\draw[big] (19.15,-1.05) -- (19.89,-1.05);

\end{tikzpicture}%
}

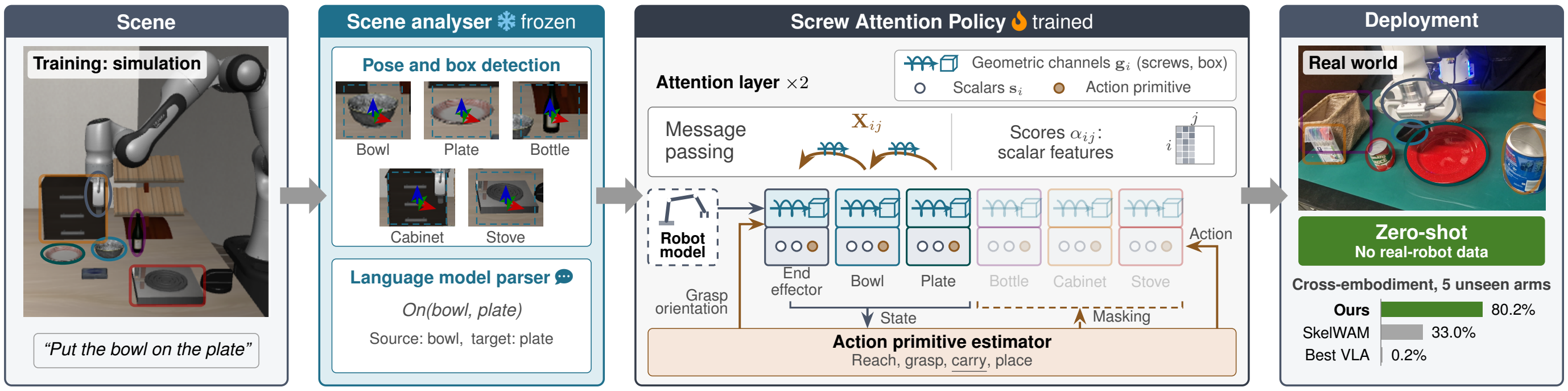
\captionof{figure}{ The Screw Attention Policy learns manipulation from geometry rather than appearance. Frozen perception and language models convert images and instructions into object poses, task goals, and geometric relations. The policy observes only the relations between the hand and scene objects to produce an end-effector twist. A 30K-parameter policy trained in simulation on a single robot transfers zero-shot to a real robot and to five unseen robot arms.}
\label{fig:architecture}
\end{center}
\vspace{2pt}

}
\makeatother

\title{Supply Geometry, Learn Execution:\\ Screw Attention for Zero-Shot Policy Transfer\\ in Robot Manipulation}

\author{\IEEEauthorblockN{Aly Magassouba}
\IEEEauthorblockA{School of Computer Science, University of Nottingham\\
Nottingham, United Kingdom\\
aly.magassouba@nottingham.ac.uk}}

\maketitle
\setcounter{figure}{1}

\begin{abstract}
We present the Screw Attention Policy, a geometry-centric manipulation framework that supplies task geometry explicitly and learns only execution. The policy operates on object poses, relative $\mathrm{SE}(3)$ transforms, and joint screw axes provided by perception and robot models. The policy requires only 30K learned parameters. Trained entirely in simulation on a single robot arm, the policy generalises to new scene layouts, transfers zero-shot across robot embodiments, and deploys directly on a real robot, without adaptation or fine-tuning. It achieves 96\% success on LIBERO, 85.8\% under layout changes, 80.2\% across five unseen robot arms, and  21 of 24 trials on a real robot. These results suggest that much of what manipulation policies learn at scale is geometric structure that can instead be supplied.
\end{abstract}

\section{Introduction}
\label{sec:intro}

Manipulation policies learned from demonstrations have mainly adopted a scaling paradigm, in which performance is improved  through larger datasets and models. This strategy has driven significant progress, but at a substantial cost in data, computational resources, and real-world  fine-tuning. Most  manipulation policies are grounded in image observations. Although this design provides rich sensory information, it also means that the geometric and physical structure required for manipulation is not directly available to the policy. The model must therefore learn to infer scene geometry, object relations, robot kinematics, and contact dynamics from pixels, in addition to learning the task-specific behaviour.

A direct consequence is a brittleness to changes in observations and embodiments. Variations in viewpoint, lighting, background, robot appearance, or kinematic structure often cause substantial degradation in performance. Sim-to-real and cross-embodiment transfer are visible manifestations of this challenge. Cross-embodiment policies are typically trained on data from many robot platforms so that embodiment-invariant behaviour can be discovered from experience \cite{ghosh2024octo,doshi2024crossformer}. Similarly, sim-to-real approaches rely on extensive domain randomisation, exposing the policy to a wide range of appearances until the real world falls within the training distribution \cite{deshpande2026molmobot}. Differently, recent vision-language-action (VLA) models have been shifting towards explicit geometric representations, including  3D foundation models \cite{lin2025evo0,li2025spatialforcing,wang2025vggt}, dedicated 3D reasoning modules \cite{xia20263dvla}, and pose-centric pretraining objectives \cite{lin2026posevla}. Taken together, these developments share a common premise: the policy spends substantial capacity learning the geometric structure of the scene and the robot.

We challenge the prevailing end-to-end paradigm in which manipulation policies are expected to infer geometry, kinematics, and dynamics from visual observations. While learning such representations is necessary when they are unavailable, much of this information can already be provided explicitly: object poses from modern perception systems~\cite{wen2024foundationpose} and kinematic structure from robot models. We therefore argue that the challenge is not how to learn geometric structure more efficiently, but whether it should be learned in the first place when it is already explicitly available. Modular systems have explored a similar decomposition~\cite{ichter2023saycan,liang2023codeaspolicies}, yet they typically learn task planning while executing hand-engineered or pre-generated skills. We instead supply geometry and learn the execution policy itself.

Given object poses and the robot model, the policy does not need to reason about appearance, viewpoint, or image statistics, and can instead operate on the geometric relations that define the task. We instantiate this idea through the Screw Attention Policy, built on Screw Attention \cite{magassouba2026screw}, a Transformer architecture designed to reason over rigid-body geometric relations.   In this framework, object and end-effector tokens carry screw representations and exchange
geometric messages through relative transforms. The policy
operates on end-effector-centric geometric relations, encoding
the scene objects relative to the end-effector rather than a specific embodiment. This design decouples task geometry from
embodiment-specific kinematics. Using this representation, a 30K-parameter policy trained solely on a single simulated arm transfers zero-shot across embodiments and to the real world, without target demonstrations or fine-tuning. Unlike existing approaches \cite{xmop2024, wu2026dexgraspzero, niu2026skelwam}, transfer is not an explicit training objective, but a consequence of our representation.

Our contributions are as follows.
\begin{itemize}
    \item We introduce the \textbf{Screw Attention Policy}, which operates on object poses and relative $\mathrm{SE}(3)$ transforms supplied by off-the-shelf perception and robot models. With \textbf{30K trainable parameters}, it achieves \textbf{96\%} on LIBERO and \textbf{85.8\%} on LIBERO-Plus under layout changes.
    \item Trained on a single simulated arm, the policy transfers  \textbf{zero-shot} to five other arms (\textbf{80.2\%}, against 33.0\% for the previous best~\cite{niu2026skelwam}) and to a real robot (\textbf{21 of 24} trials).
    \item We show that supplied poses alone are not sufficient: baselines given the same object poses remain below \textbf{30\%} success on rearranged layouts, whereas our policy reaches \textbf{74.7\%}. The same policy further exhibits compositional behaviour on unseen object pairings and recovers from failed grasps without explicit recovery training.
    
\end{itemize}

\section{Method}
\label{sec:method}

\subsection{Preliminaries on Screw Attention}
\label{sec:prelim}
\subsubsection{Motivation}
Consider the task of grasping a bowl. To decide the next motion, a robot needs to know where the bowl is relative to the hand: its distance, direction, and orientation. If the entire scene is translated or rotated, these relations remain unchanged, and so does the desired motion. Manipulation can therefore be viewed as reasoning over geometric relations between bodies, rather than directly over pixels.

Screws provide a natural representation for such relations~\cite{featherstone2008}. A screw is a six-dimensional vector that combines angular and linear components in the coordinate frame of a body. Poses, relative displacements, end-effector motions, and articulation constraints can all be represented in this form. For example, the displacement from the hand to the bowl is represented as the screw $[\mathbf{0}; \mathbf{p}]$. The commanded hand motion is another type of screw, a twist $[\boldsymbol{\omega}; \mathbf{v}]$, which combines angular and linear velocity.  This shared representation enables the policy to reason about objects, robot states, and their interactions with the same underlying operations.

\subsubsection{Core principle}
Manipulation requires combining geometric quantities attached to different bodies.  Let $T_i$ and $T_j$ denote the poses of bodies $i$ and $j$, and let $T_{ij}=T_i^{-1}T_j=(\mathbf{R},\mathbf{p})$ be their relative pose with the rotation matrix $\mathbf{R}$. The associated spatial transform \[ \X_{ij}= \begin{bmatrix} \mathbf{R} & \mathbf{0}\\ [\mathbf{p}]_\times\mathbf{R} & \mathbf{R} \end{bmatrix} \] re-expresses any screw from the frame of $j$ into the frame of $i$. Screws expressed in different frames can then be compared and combined within a common coordinate system. Consequently, the representation is independent of how coordinate frames are attached to robots, grippers, or objects. Changing a frame modifies both the screws and transforms consistently while preserving the underlying geometry. Screw Attention~\cite{magassouba2026screw} exploits this property by exchanging screw messages through relative transforms, forming the basis of our manipulation policy.

\subsubsection{Screw Attention}
In Screw Attention, each body is represented by a Transformer token associated with a pose $T_i\in\mathrm{SE}(3)$. Each ordered pair of tokens $(i,j)$ carries the relative transform $\X_{ij}$. Additionally, articulated bodies  carry a joint screw $\mathbf{S}_j$ encoding the type and axis of the joint (Section~\ref{sec:tokens}). Neither $\X_{ij}$ nor $\mathbf{S}_j$ is learned. Both are supplied by the robot model and object pose estimates.

Each token encodes two feature streams. The scalar features
$\mathbf{s}_i \in \mathbb{R}^d$ are frame-independent quantities, such as distances. The geometric features $\mathbf{g}_i\in\mathbb{R}^{k\times6}$ contain $k$ screw-valued channels expressed in the local frame of token $i$, encoding quantities such as object directions, gravity, or joint axes.

The tokens are processed by a subsequent attention layer. The attention weight $\alpha_{ij}$ between tokens $i$ and $j$ is given by 
\begin{equation}
\alpha_{ij} = \operatorname{softmax}_{j}\Big(
\tfrac{1}{\sqrt{d_h}}(\mathbf{W}_q\mathbf{s}_i)^{\top}\mathbf{W}_k\mathbf{s}_j
+ \mathbf{w}^{\top}\boldsymbol{\psi}_{ij}\Big),
\label{eq:scores}
\end{equation}
where $d_h$ is the attention-head dimension, $\mathbf{W}_q$ and $\mathbf{W}_k$ are learned query and key projections, and $\mathbf{w}$ is a learned weight vector. The first term  in Eq.~\ref{eq:scores} is the standard attention score computed from scalar features. The additional term $\boldsymbol{\psi}_{ij}$ captures the geometric relationship between bodies $i$ and $j$. It includes features derived from their relative pose, such as distance and axis alignment, together with frame-invariant pairings between their geometric channels after transport into a common frame by $\X_{ij}$. These pairings are computed using the Killing and Klein bilinear forms on screws~\cite{selig2005geometric}. The resulting attention depends on both the relative configuration of bodies and the alignment of their geometric features.

Both the scalar and geometric streams are updated using standard attention operations. For the geometric stream, messages are first transported into the receiver's frame and then aggregated as, 
\begin{equation}
\mathbf{g}_i \leftarrow \mathbf{g}_i + \boldsymbol{\gamma}(\mathbf{s}_i) \odot
\mathbf{W}_{go}\Big[\textstyle\sum_j \alpha_{ij}\, \X_{ij}\,(\mathbf{W}_g\,
\mathbf{g}_j)\Big],
\label{eq:update}
\end{equation}
where $\mathbf{W}_g$ and $\mathbf{W}_{go}$ mix only the $k$ geometric channels, and $\boldsymbol{\gamma}(\mathbf{s}_i)$ is a channel-wise gate computed from the scalar features. Crucially, no learned parameter acts on the six spatial components of a screw. The six spatial components are transformed only by $\X_{ij}$ when a screw is transported between frames.
As shown in~\cite{magassouba2026screw}, this design makes the layer equivariant to arbitrary changes of reference frame. 

In this work, we use a simplified instantiation of Screw Attention. Unlike the original formulation, which represents every robot link as a token, we represent the robot only through an end-effector token. This end-effector-centric representation removes embodiment-specific information while preserving the task geometry required for manipulation.

\subsection{System overview}
\label{sec:system}
Our architecture is shown in Fig.~\ref{fig:architecture}. The system consists of two stages: a scene analyser that converts raw sensory observations into a geometric scene representation, and a geometry-centric policy that generates actions from this representation.

The scene analyser combines off-the-shelf perception~\cite{wen2024foundationpose} and a language model. Using wrist-mounted and external cameras, the perception module identifies scene objects and estimates their poses and bounding boxes. Alongside the perception module, the language model parses the instruction into a sequence of steps, each specified by an action, a target object, and, when relevant, a destination. To resolve referring expressions and ambiguous instructions, the language model additionally receives a symbolic scene description containing object labels and estimated poses. The resulting geometric scene description, together with the parsed instruction, is provided to the Screw Attention Policy.  Our policy is agnostic to the particular scene-analysis pipeline and can interface with any system that provides these quantities.

The policy consists of two components: (i) a Screw Attention Transformer which encodes the scene with one token per object and an end-effector token, and (ii)  an action-primitive estimator which infers from the scene state the primitive to execute and the desired grasp and release orientations in the object frame.  Conditioned on these features, the policy predicts an end-effector twist, which is executed by the robot.

A key property of this architecture is the separation between perception and control. The policy never processes images directly. Instead, it operates exclusively on object poses and geometric relations. As a result, camera viewpoint, illumination, texture, and background influence the policy only through the scene analyser. The control policy is therefore invariant, by construction, to photometric variations and camera configurations. Unlike pixel-based approaches, which must jointly learn perception and manipulation, our approach confines the visual component of the sim-to-real gap to the perception stack.

\begin{table}[t]
\centering
\caption{Token contents. Geometric channels are screws.}
\label{tab:tokens}
\footnotesize
\setlength{\tabcolsep}{3pt}
\renewcommand{\arraystretch}{1.15}
\newcommand{\cell}[1]{\begin{tabular}[t]{@{}l@{}}#1\end{tabular}}
\begin{tabular}{@{}lll@{}}
\toprule
Token & Scalars & Geometric channels \\
\midrule
End effector & gripper opening
             & \cell{hand $\to$ source;\\
                     source $\to$ destination;\\
                     grasp, release orientation;\\
                     gravity} \\[2pt]
Scene token  & \cell{hand distance, height;\\
                     role;\\
                     class (object);\\
                     joint position (part)}
             & \cell{own axis, or $\mathbf{S}_j$ (part);\\
                     box (object, part);\\
                     gravity} \\
\midrule
All tokens   & action primitive & -- \\
\bottomrule
\end{tabular}
\end{table}

\subsection{Tokens and relations}
\label{sec:tokens}
A task is represented as a set of scene tokens together with an end-effector token. Scene tokens correspond to objects, articulated parts, and goal regions. The scalar and geometric features carried by each token are summarised in Table~\ref{tab:tokens}.

Scene tokens encode the geometric structure of the environment. Rigid objects are represented by geometric descriptors (bounding boxes, axis, gravity direction) attached to their local frame, while articulated parts additionally carry the joint screw $\mathbf{S}_j$, enabling the policy to reason about task-relevant kinematic constraints. Articulated fixtures are decomposed into one token per movable component. For example, a cabinet with three drawers is represented by three drawer tokens, each associated with its corresponding prismatic joint. 

The end-effector token encodes the task geometry relative to the hand. Its geometric stream contains the hand-to-source displacement, the source-to-destination displacement, and the desired grasp and release orientations. The grasp and release orientations are predicted by the action-primitive estimator, which is trained from contact events in the demonstrations.

Finally, all tokens receive the predicted action primitive through their scalar stream. The action-primitive estimator also assigns task-specific roles, such as source or destination. These roles restrict the attention of the end-effector token to the scene tokens involved in the current manipulation step, allowing a single policy to focus on the objects relevant to the task.

\subsection{Action-primitive estimator}
\label{sec:verb}
Manipulation tasks are naturally structured as a sequence of interaction stages. For example, placing an object typically involves reaching, grasping, carrying, and placing. While a sufficiently large policy could learn these stages implicitly, doing so forces it to discover task structure from correlations in the training data. This increases the risk of overfitting to embodiment- or dataset-specific cues rather than learning the semantics of the interaction. We instead expose this structure explicitly through action primitives.
The action-primitive estimator predicts, from the scene state, when the interaction moves from one primitive to the next, among \emph{reach}, \emph{grasp}, \emph{carry}, \emph{place}, \emph{pull}, \emph{push}, and \emph{press}. The current primitive is injected into all tokens through the scalar stream and conditions the policy on the type of interaction being executed without prescribing the motion. 

\subsection{Output}
\label{sec:output}

The policy predicts a short horizon of end-effector actions at each forward pass. Each action consists of a twist expressed in the end-effector frame and a gripper command. The horizon is represented as a chunk of future actions, following the strategy introduced in ACT~\cite{zhao2023act}. Each predicted twist therefore retains the embodiment-independent properties of the policy representation. At execution, overlapping chunks are averaged with weights that decay with the age of the prediction, which smooths the commanded motion. The resulting twist is then executed by the robot's operational-space controller.

The policy maps task geometry directly to end-effector motions expressed in the hand frame. As a result, deployment on a new robot requires only mapping the normalised gripper command to the gripper's opening range.

\section{Experiments}
\label{sec:experiments}

\subsection{Protocol}
\label{sec:protocol}
The policy is trained by behaviour cloning on the 2,000 human demonstrations of LIBERO~\cite{libero2023} (50 per task). The complete system, including the Screw Attention Policy and action-primitive estimator, contains approximately 30K trainable parameters and trains in about two hours on a single L4 GPU. No pretraining, generated data, demonstrations from other robots, real-world data, or fine-tuning are used.

We evaluate a single policy on the four LIBERO suites, using the standard LIBERO evaluation protocol with 50 episodes per task. Robustness is measured on LIBERO-Plus~\cite{liberoplus2025}, which perturbs object layouts and robot initial states. Cross-embodiment transfer is evaluated on LIBERO-Cross10~\cite{niu2026skelwam}, where the same policy is deployed without retraining on five robot arms using the benchmark's operational-space controller.

Unless stated otherwise, object poses are estimated using CNOS~\cite{nguyen2023cnos} and FoundationPose~\cite{wen2024foundationpose} from object meshes, readily available from the training simulator, though the policy itself is agnostic to the 6D  pose estimator.  Rows labelled \emph{True Pose} replace estimated poses with simulator ground truth to isolate the effect of perception errors. Reported parameter counts include all trainable components of our method; perception and language models are frozen off-the-shelf systems.

\subsection{LIBERO and LIBERO-Plus}
\label{sec:libero}
\begin{table}[t]
\centering
\caption{Success (\%) on LIBERO and LIBERO-Plus~\cite{liberoplus2025}.}
\label{tab:canonical}
\footnotesize
\setlength{\tabcolsep}{1.4pt}
\renewcommand{\arraystretch}{1.1}
\begin{tabular}{@{}lrccccccc@{}}
\toprule
 & & \multicolumn{5}{c}{Canonical} & \multicolumn{2}{c}{LIBERO-Plus} \\
\cmidrule(lr){3-7}\cmidrule(lr){8-9}
Method & Trained & Spatial & Object & Goal & Long & Mean & Robot & Layout \\
\midrule
OpenVLA \cite{kim2024openvla}         & 7B   & 84.7 & 88.4 & 79.2 & 53.7 & 76.5 & 4.1 & 31.6 \\
$\pi_0$ \cite{black2025pi0}           & 3.3B & 96.8 & 98.8 & 95.8 & 85.2 & 94.2 & 6.6 & 70.4 \\
$\pi_{0.5}$ \cite{black2025pi05}      & 2.3B & 98.8 & 98.2 & \best{98.0} & 92.4 & 96.9 & 76.6 & 86.0 \\
$\pi_{0.5}$+3DVLA \cite{xia20263dvla} & 2.4B & --   & --   & --   & --   & --   & 77.4 & 86.6 \\
OpenVLA-OFT \cite{kim2025openvlaoft}  & 7.5B & 97.6 & 98.4 & 97.9 & \best{94.5} & \best{97.1} & 37.2 & 77.1 \\
SmolVLA \cite{shukor2025smolvla}      & 450M & 90.0 & 96.0 & 92.0 & 71.0 & 87.3 & -- & -- \\
Pose-VLA \cite{lin2026posevla}        & 3B   & 96.5 & 98.0 & 97.1 & 92.4 & 96.0 & -- & -- \\
\midrule
\ours Ours (True Pose) & 30K  & 99.0 & 99.4 & 96.2 & 93.4 & 97.0 & \best{83.6} & \best{88.2} \\
\ours Ours          & 30K  & \best{99.2} & \best{99.6} & 91.0 & 94.2 & 96.0 & 78.2 & 85.8 \\
\bottomrule
\end{tabular}
\end{table}

Table~\ref{tab:canonical} compares the proposed policy against recent VLA models on the four LIBERO suites. Despite containing only 30K learned parameters, the Screw Attention Policy achieves 96\% average success, comparable to models several orders of magnitude larger. Using true poses increases the average only marginally, from 96.0\% to 97.0\%, so the system does not depend on precise pose estimation. The main perception gap appears on the Goal suite (91.0\% versus 96.2\%).

The Screw Attention Policy maintains its performance under LIBERO-Plus perturbations~\cite{liberoplus2025}, achieving 78.2\% success under robot initialisation perturbations and 85.8\% under layout perturbations. These results are comparable to the strongest VLA baselines despite a parameter count that is five orders of magnitude smaller. With ground truth poses, our approach exceeds all reported baselines. These results support the central hypothesis of this work: supplying geometry explicitly, in a representation that preserves it, enables compact policies to generalise.

\subsection{Zero-shot transfer across arms}
\label{sec:crossemb}
\ifdefined\combinedtables\else\begin{table}[t]\fi
\centering
\caption{Zero-shot transfer success (\%) on LIBERO-Cross10~\cite{niu2026skelwam}.}
\label{tab:crossemb}
\footnotesize
\renewcommand{\arraystretch}{1.1}
\setlength{\tabcolsep}{2.5pt}
\begin{tabular}{lcccccc}
\toprule
 & \multicolumn{5}{c}{Target arms, zero-shot} & \\
\cmidrule(lr){2-6}
Method & UR5e & Kinova3 & IIWA & UR10e & xArm7 & Mean \\
Robotiq gripper (ours) & 85 & 85 & 140 & 85 & 85 & \\
\midrule
\cat{7}{Imitation learning} \\
Diffusion Policy \cite{chi2023diffusion} & 2.0 & 0.0 & 5.0 & 3.0 & 3.0 & 2.6 \\
\cat{7}{Vision-language-action models} \\
$\pi_{0.5}$ \cite{black2025pi05}        & 0.0 & 0.0 & 0.0 & 0.0 & 0.0 & 0.0 \\
OpenVLA-OFT \cite{kim2025openvlaoft}    & 0.0 & 0.0 & 0.0 & 0.0 & 1.0 & 0.2 \\
\cat{7}{World action models} \\
FastWAM \cite{yuan2026fastwam}          & 0.0 & 0.0 & 0.0 & 0.0 & 4.0 & 0.8 \\
LaWAM \cite{chen2026lawam}              & 0.0 & 4.0 & 0.0 & 1.0 & 2.0 & 1.4 \\
\cat{7}{Cross-embodiment methods} \\
Mirage \cite{chen2024mirage}            & 0.0 & 0.0 & 0.0 & 0.0 & 4.0 & 0.8 \\
RoVi-Aug \cite{chen2024roviaug}         & 15.0 & 7.0 & 9.0 & 4.0 & 13.0 & 9.6 \\
SkelWAM \cite{niu2026skelwam}           & 13.0 & 49.0 & 22.0 & 62.0 & 19.0 & 33.0 \\
\midrule
\cat{7}{Supplied geometry} \\
\ours Screw Attention Policy & \best{81.0} & \best{87.0} & \best{85.0} & \best{71.0} & \best{77.0} & \best{80.2} \\
\bottomrule
\end{tabular}
\ifdefined\combinedtables\else\end{table}\fi

Table~\ref{tab:crossemb} evaluates zero-shot cross-embodiment transfer on LIBERO-Cross10~\cite{niu2026skelwam}. The policy is deployed without adaptation on five robot manipulators equipped with different grippers. Baselines are reported as in~\cite{niu2026skelwam} and averaged over the same five target arms. We follow the published protocol, with object poses estimated from simulated camera images. Existing manipulation policies transfer poorly across embodiments. General-purpose VLAs and world-action models achieve near-zero success, while methods explicitly designed for cross-embodiment transfer reach at most 33.0\% on average. In contrast, the Screw Attention Policy achieves 80.2\% average success, more than doubling the previous best result. Performance remains consistently high across all target robots, ranging from 71.0\% to 87.0\%.

These results suggest that task geometry transfers more readily than robot-specific representations. The policy receives no joint states and no embodiment-specific features. It reasons over geometric relations expressed relative to the end effector, represented by object poses, relative transforms, and task constraints.

\subsection{Insight into the geometry}
\label{sec:geometry}
\ifdefined\combinedtables\else\begin{table}[t]\fi
\centering
\caption{Success (\%) on LIBERO, under the LIBERO-PRO position swap~\cite{liberopro2025},
after a forced failure, and on a new task: the bowl placed into the basket}
\label{tab:geometry}
\footnotesize
\setlength{\tabcolsep}{2.6pt}
\renewcommand{\arraystretch}{1.1}
\begin{tabular}{@{}lccccc@{}}
\toprule
 & LIBERO & LIBERO-PRO & \multicolumn{2}{c}{Failure recovery} & New \\
\cmidrule(lr){2-2}\cmidrule(lr){3-3}\cmidrule(lr){4-5}\cmidrule(lr){6-6}
Method & Canonical & Swap & Missed & Dropped & task \\
\midrule

MLP (1M)                     & 92.0 & 0.3  & 0.0  & 8.0  & 0.0 \\
Transformer (1M)             & 89.3 & 15.8 & 16.0 & 4.0  & 0.0 \\
Transformer (matched)        & 93.5 & 13.5 & 12.0 & 0.0  & 0.0 \\
\quad + attention mask       & 92.8 & 27.7 & 0.0  & 20.0 & 0.0 \\
\midrule
\ours Screw Attention Policy    & \best{98.5} & \best{74.7} & \best{76.0} & \best{56.0} & \best{100.0} \\
\bottomrule
\end{tabular}
\ifdefined\combinedtables\else\end{table}\fi

Table~\ref{tab:geometry} investigates whether the policy learns task geometry or merely reproduces demonstrated trajectories. All baselines receive the same observations as the proposed policy, including object poses and action primitives. The only difference is the representation. 

To decouple perception from execution, all methods are evaluated with ground-truth poses on 600 episodes from the Spatial, Goal, and Object suites. On the canonical benchmark, all methods achieve comparable performance (89--99\%), indicating that standard evaluations are insufficient to distinguish between fundamentally different solutions. The distinction appears only when the state deviates from the demonstrations.

Under the LIBERO-PRO position swap, the proposed policy achieves 74.7\% success, whereas all baselines remain below 30\%. A pose-based approach is therefore insufficient: the 1M-parameter MLP collapses despite receiving the same object poses. Similarly, a matched-size Transformer with the same inputs performs poorly. Increasing the model capacity or adding  an attention mask yields only modest improvements, suggesting that generalisation stems from our representation.

The same trend appears on a previously unseen task, \emph{Place the bowl in the basket}. The bowl and the basket both appear in training, but never together. The Screw Attention Policy succeeds in all trials, whereas all baselines fail completely. The policy adapts its trajectory to the geometry of the basket, lifting the bowl above the rim before placing, suggesting that previously learned manipulation skills are composed in a novel geometric setting rather than recalled from demonstrations.

Finally, despite never being trained for recovery, the policy completes 76.0\% of missed-grasp trials and 56.0\% of dropped-object trials by regrasping from the current scene state. Crucially, this recovery does not rely on the action-primitive estimator switching back to a grasping phase; even when the primitive remains fixed, the policy re-approaches and regrasps the object. Because actions are driven directly by relative transforms, a missed grasp or drop immediately alters the geometric relations and induces corrective motion. Baselines instead tend to continue the demonstrated sequence without re-engaging with the object. Robustness to layout changes, compositional generalisation, and failure recovery emerge from the geometric structure of the representation rather than from model scale or explicit supervision. Hand-centred geometry removes dependence on cameras and robot embodiments, whereas the screw representation ties the motion to the current geometric relations.

\subsection{Real robots}
\label{sec:real}
\newcommand{\realstrip}[1]{%
  \begin{tikzpicture}[inner sep=0, outer sep=0]
    \node[anchor=north west] (t1) at (0,0) {\includegraphics[width=\realtilewidth]{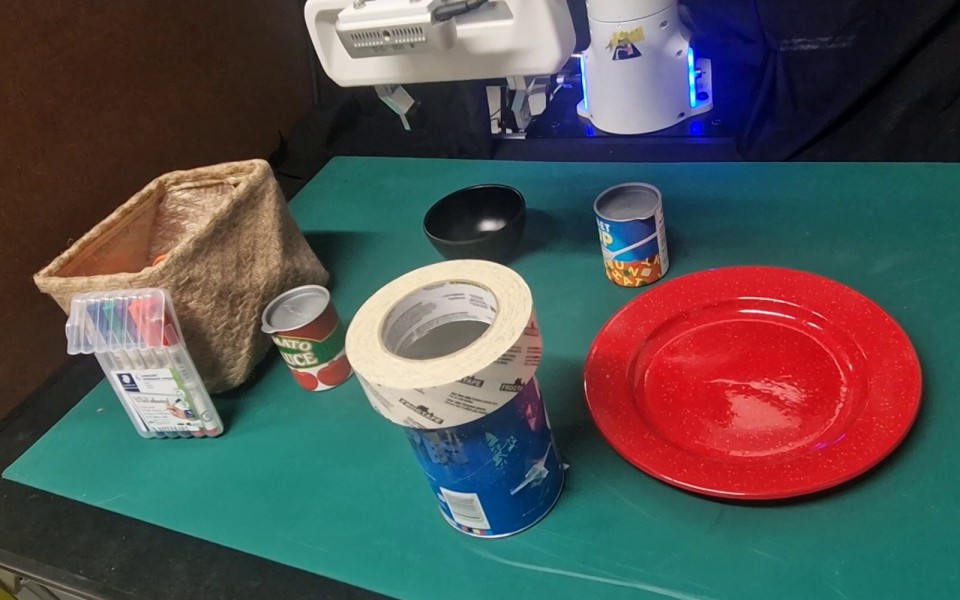}};
    \node[anchor=north west] (t2) at ([xshift=\realgap]t1.north east) {\includegraphics[width=\realtilewidth]{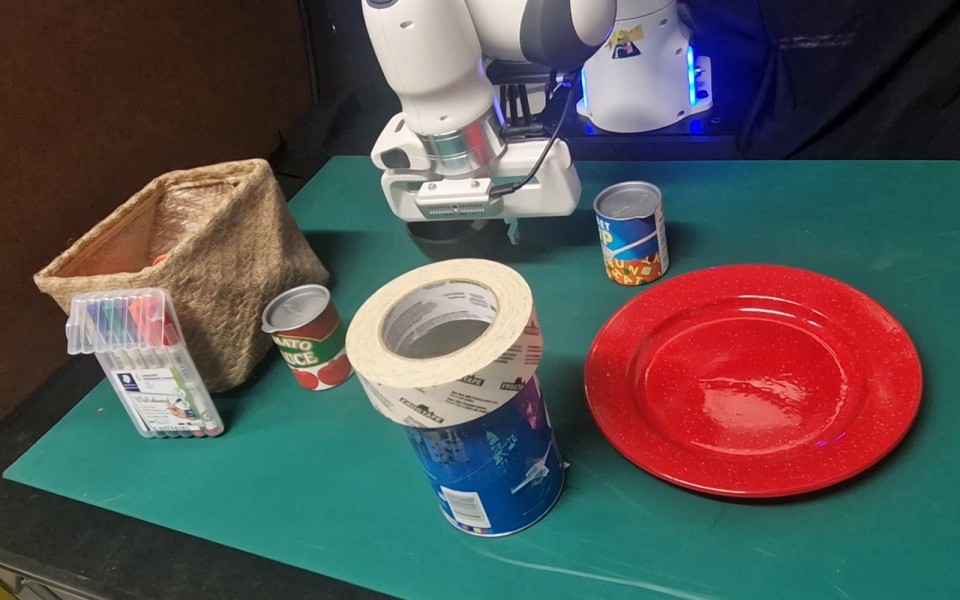}};
    \ifnum#1=2
      \node[anchor=north west] (t3) at ([yshift=-\realgap]t1.south west) {\includegraphics[width=\realtilewidth]{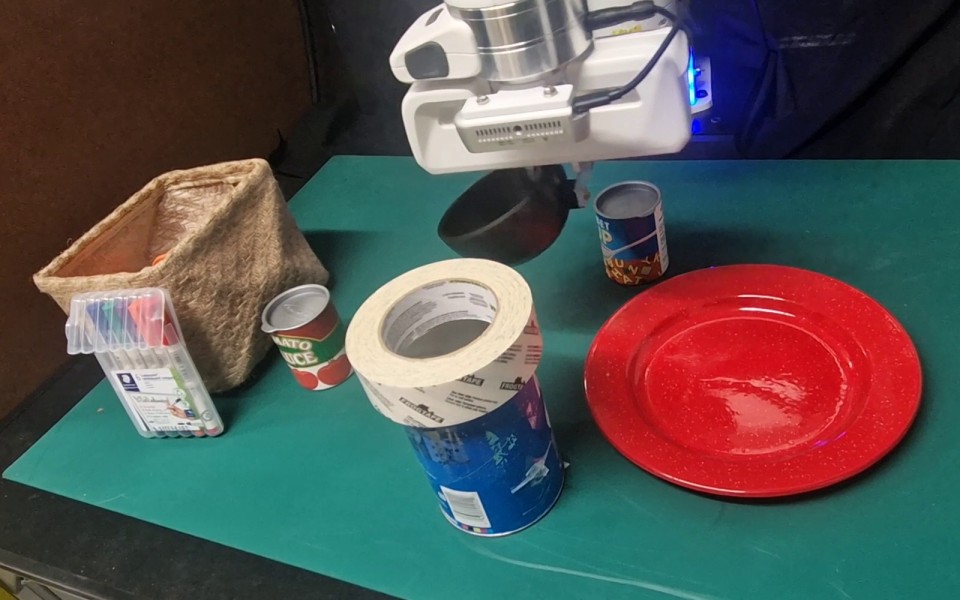}};
    \else
      \node[anchor=north west] (t3) at ([xshift=\realgap]t2.north east) {\includegraphics[width=\realtilewidth]{img/real_t05_3.jpg}};
    \fi
    \node[anchor=north west] (t4) at ([xshift=\realgap]t3.north east) {\includegraphics[width=\realtilewidth]{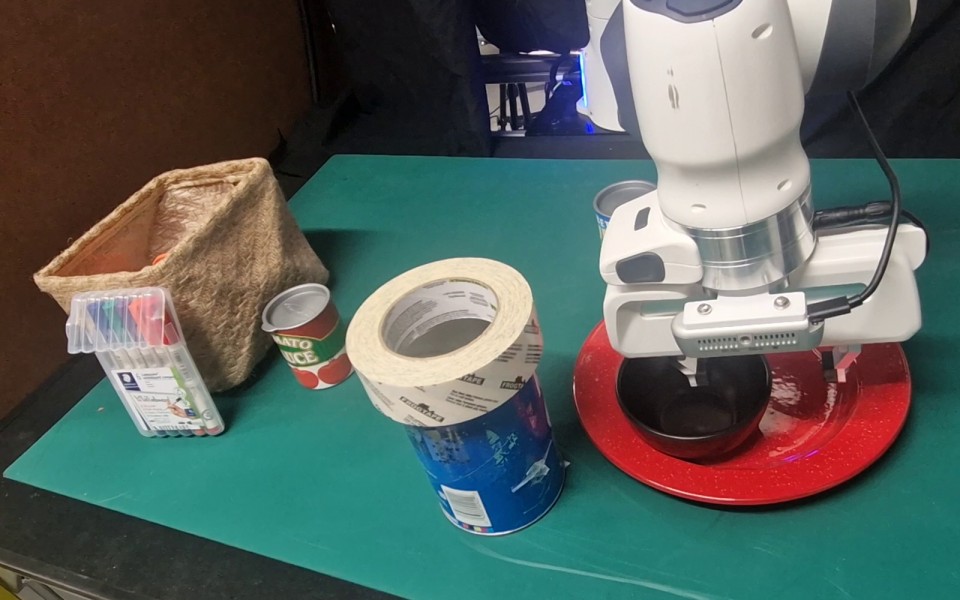}};
    \foreach \n/\t in {t1/1 Approach, t2/2 Grasp, t3/3 Carry, t4/4 Place}
      \node[anchor=north west, fill=white, fill opacity=0.8, text opacity=1,
            inner sep=1.2pt, font=\realtilefont] at (\n.north west) {\t};
    \ifnum#1=2 \def\realleft{t3}\else\def\realleft{t1}\fi
    \filldraw[fill=white, draw=black, line width=0.5pt] ([xshift=0.25pt, yshift=0.25pt]\realleft.south west) rectangle ([xshift=-0.25pt, yshift=\realbannerheight]t4.south east);
    \path (\realleft.south west) -- ([yshift=\realbannerheight]t4.south east)
      node[midway, text=black, font=\realcountfont\mdseries]
      {{\realcountfont Success rate on two Pandas 87.5\%} (9/12 and 12/12)};
  \end{tikzpicture}}
\ifdefined\combinedtables
  \def\realtilewidth{0.2425\linewidth}\def\realgap{0.01\linewidth}\def\realtilefont{\tiny\sffamily}%
  \def\realcountfont{\scriptsize\sffamily\bfseries}\def\realbannerheight{2.6ex}%
  \realstrip{4}
  \captionof{figure}{The policy on a real Panda, trained only in simulation, executing
\emph{``Put the bowl on the plate''}. Over 12 runs
on random layouts, 9 succeeded, two
of them after a missed grasp and a retry.}
  \label{fig:real}
\else
  \def\realtilewidth{0.492\columnwidth}\def\realgap{0.016\columnwidth}\def\realtilefont{\scriptsize\sffamily}%
  \def\realcountfont{\small\sffamily\bfseries}\def\realbannerheight{3.0ex}%
  \begin{figure}[t]
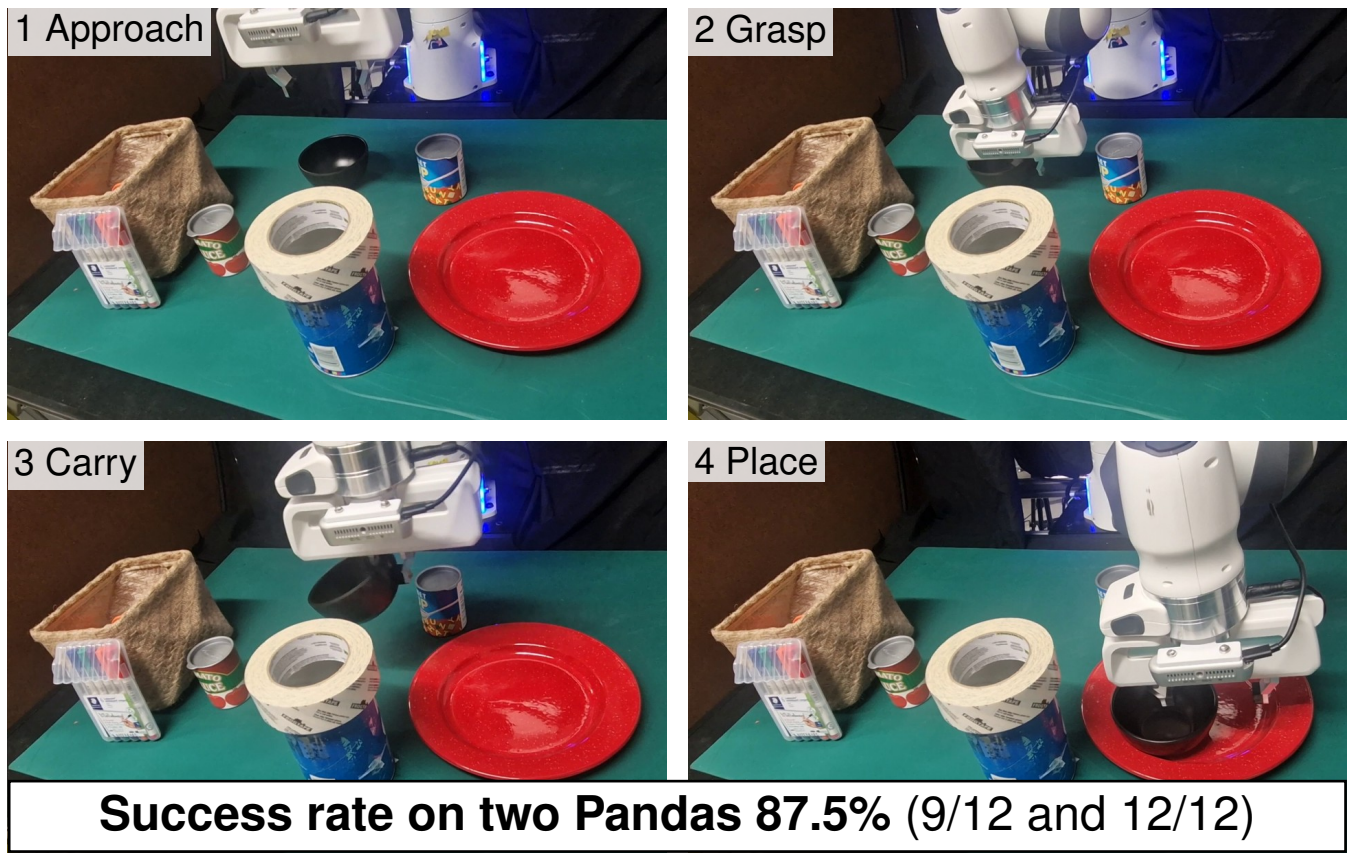

  \centering
  \realstrip{2}
  \caption{The policy on a real Panda, trained only in simulation, executing
\emph{``Put the bowl on the plate''}. Over 24 runs on different random layouts, 21 succeeded, two
of them after a missed grasp and a retry.}
  \label{fig:real}
  \end{figure}
\fi

We performed a preliminary real-world validation on two Franka Panda using the same policy trained entirely in simulation. The deployment pipeline is identical to the simulated experiments and provides the same geometric scene representation to the policy. On a laptop CPU, the policy requires 9\, ms per control step, against a 50 ms control period.

As illustrated in Fig.~\ref{fig:real}, the policy transfers zero-shot from simulation to the real robot. We evaluated it over two sessions of 12 trials each, using scene layouts substantially different from those observed in simulation. The policy succeeded in 21 out of 24 trials (87.5\%) without real-world demonstrations. The experiments also provide preliminary evidence that the recovery behaviours observed in simulation transfer to the real world. Two successful executions recovered from an initial missed grasp and subsequently completed the task. 
These experiments are intended as a feasibility study. In the full paper, a more extensive evaluation will be conducted on multiple tasks, embodiments, and scenes.

\section{Conclusion}
\label{sec:conclusion}
This paper introduced the Screw Attention Policy, a geometry-centric manipulation framework that separates perception from control and learns execution directly from object poses and relative transforms. Despite containing only 30K parameters and training on a single simulated robot, the policy achieves strong performance on LIBERO, robust generalisation under scene perturbations, and zero-shot transfer across embodiments and to the real world. The resulting policy also exhibits compositional and recovery behaviours without explicit supervision. Together, these results suggest that many capabilities commonly associated with scale can instead emerge from an appropriate geometric representation. Supplying geometry explicitly offers a practical alternative to learning scene structure from pixels, enabling compact policies that generalise beyond the configurations observed during training.

\bibliographystyle{plainnat-short}
\bibliography{refs}

\end{document}